\documentclass[letterpaper]{article} 
\usepackage{aaai2027}
\usepackage[hyphens]{url}  
\usepackage{graphicx} 
\usepackage[numbers]{natbib}  
\usepackage{caption} 
\usepackage{algorithm}
\usepackage{algorithmic}
\usepackage{booktabs}
\usepackage{amsmath}
\DeclareMathOperator*{\argmin}{arg\,min}
\usepackage{amssymb}
\usepackage{xcolor}
\definecolor{rxred}{RGB}{0,0,0}
\newcommand{\rxcolor}{rxred}

\newenvironment{rxblock}{\begingroup\color{\rxcolor}}{\endgroup}

\usepackage{multirow}
\usepackage{makecell}

\title{AURA: Active-Response Attribution under Treatment Ambiguity in Bacterial
Cytological Profiling}

\author{
    Kartik Jhawar\textsuperscript{\rm 1, 2}*,
    Mrunmayee Deshpande\textsuperscript{\rm 1},
    Wilfried Moreira\textsuperscript{\rm 1},
    Guillermo C. Bazan\textsuperscript{\rm 1,3},
    Lipo Wang\textsuperscript{\rm 1, 2}
}
\affiliations{
    \textsuperscript{\rm 1}Institute for Digital Molecular Analytics and Science, Nanyang Technological University, Singapore 636921 \\
    \textsuperscript{\rm 2}School of Electrical and Electronic Engineering, Nanyang Technological University, Singapore 639798 \\
    \textsuperscript{\rm 3}Institute for Functional Intelligent Materials, National University of Singapore, Singapore 117544\\
    *kartikvi001@e.ntu.edu.sg
}

\begin{document}
\nocopyright

\maketitle

\begin{abstract}
When a bacterial sample is exposed to several antibiotics, not every applied drug necessarily acts: if the organism is resistant to one of them, that drug leaves no morphological trace. The clinically meaningful quantity is therefore not which antibiotics were \emph{applied}, but which ones were \emph{active}. We show that these two are sharply decoupled in real \textit{E.\,coli} microscopy---naively assuming the applied combination equals the active one is correct only about 37\% of the time---yet existing computational tools are ill-suited to recovering the active set. Forward perturbation models such as scGen, CPA, and IMPA are designed to predict appearance from treatment, not the reverse, and inverting them degrades sharply; discriminative image classifiers tend to memorise strain- and batch-specific texture and fail to transfer across experimental replicates. We introduce AURA, which reframes the task as constrained, energy-based inverse attribution. Its central inductive bias is that the active set must be a subset of the applied set; this collapses the candidate space and lets AURA infer the active subset of applied antibiotics by decomposing residual morphology into antibiotic response atoms and selecting the subset with the lowest reconstruction energy, using no strain label at test time. AURA-E adds evidence-aware abstention, withholding a prediction when candidate explanations remain near-equally plausible. On cross-replicate transfer in an E. coli cytological profiling dataset, AURA recovers the active antibiotic combination with 95.47\% exact-match accuracy. This confirms a substantial decoupling from the naive applied-treatment rule (37.43\%) and improves upon the strongest non-AURA baseline by 3.41 percentage points. By adding selective abstention, AURA-E reaches 97.86\% exact match at 80\% coverage. Furthermore, a public-data stress test on a frozen RxRx3-derived pseudo-cocktail benchmark indicates the active-subset principle extends beyond bacteria, improving active-pair recovery by 36.68 percentage points over the strongest baseline. Ultimately, these results support AURA as a constraint-aware inference approach for active-response attribution, succeeding where generic multi-label classification fails.
\end{abstract}

\section{Introduction}
Choosing the right antibiotic is a time-critical clinical decision. Classical susceptibility assays---minimum inhibitory concentration testing and disk diffusion---answer it reliably but take one to two days, require specialised handling, and scale poorly to combinations of drugs. High-content microscopy offers a faster route. Because different antibiotics disrupt different cellular machinery, a treated bacterium changes shape in characteristic ways: ciprofloxacin elongates cells by inhibiting DNA replication, $\beta$-lactams such as ceftriaxone distort or lyse the cell wall, and aminoglycosides such as gentamicin perturb protein synthesis \cite{drlica1997,spratt1975,typas2012}. Reading a drug's effect from this morphology, known as \emph{bacterial cytological profiling} (BCP) \cite{nonejuie2013,zoffmann2019}, can in principle report drug activity within hours rather than days.

The difficult point is that the treatment placed in the dish is not necessarily the response expressed in the image. If ciprofloxacin, ceftriaxone, and gentamicin are applied together, a sensitive wild-type strain may show all three active responses, whereas a resistant strain may show only one, or none. Thus, the applied antibiotic set is known from the experiment, but the active response subset is hidden in the morphology. This treatment--response decoupling is the central problem addressed in this work. It is different from ordinary treatment classification: the goal is not to map an image directly to 000'', 010'', or ``111'', but to determine which applied antibiotics left an observable response signature after accounting for baseline morphology. The genuine task is the inverse one: given the observed cells and the known applied set, infer which of the applied drugs actually acted.

This inverse problem is hard for three reasons. First, when two drugs are simultaneously active their phenotypes blend, and the mixture cannot be separated by visual inspection. Second, the same drug produces somewhat different morphology on different strains and under different imaging conditions. Third, and most damaging for machine learning, experiments are run as independent biological replicates: a model that latches onto the appearance of one replicate---its background, cell density, or staining---will fail when shown a new one. Robust attribution must therefore learn the relationship between phenotype and drug \emph{action} rather than the look of a particular batch.

Existing computational approaches do not directly solve this inverse active-subset problem. Standard discriminative classifiers can learn mappings from images to treatment or response labels, but they do not explicitly encode the fact that the active response must be a subset of the applied treatment. They also risk learning replicate-, strain-, or acquisition-specific appearance rather than transferable antibiotic-response evidence \cite{dixit2016}. Conversely, forward perturbation models---scGen \cite{lotfollahi2019}, CPA \cite{lotfollahi2023}, and the image-based IMPA \cite{palma2025}, together with more recent generative successors \cite{wang2025,demirel2025,klein2025}---learn to predict how cells would look \emph{after} a given treatment. They are powerful generators, but identifying the treatment from the image is not their
design goal, and inverting them (searching for the treatment whose predicted appearance best matches the observation) degrades substantially, as we show. The same issue arises for image-based forward morphology models: they address what would the cell look like if perturbed?'' rather than which response components explain this observed morphology?''.

We propose AURA, which reframes antibiotic attribution problem as constrained, energy-based inverse inference. Given an observed microscopy field and the known applied antibiotic set, AURA enumerates biologically valid active subsets and scores each candidate by how well its learned response atoms reconstruct the observed residual morphology. In simple terms, the model asks whether the cells look more like baseline plus a ciprofloxacin response, baseline plus a ceftriaxone response, baseline plus both, or baseline with no active response. The winning explanation is the active subset with the lowest reconstruction energy. This applied-set constraint is central: a drug that was not applied cannot be active, but a drug that was applied may still be inactive if the bacterium resists it. To ensure the model learns true biological responses rather than exploiting visual artifacts, AURA deliberately uses \emph{frozen} ImageNet-pretrained ResNet-18 \cite{he2016}, ViT \cite{dosovitskiy2021}, and DINOv2 \cite{oquab2024} encoders to prevent the model from retraining its way into memorising batch-specific appearance. We further introduce AURA-E, an evidence-aware selective prediction layer. In some images, two candidate active subsets may explain the morphology almost equally well. Rather than forcing a brittle prediction, AURA-E uses the energy landscape to abstain on uncertain cases. This is particularly important for biomedical use, where a system that knows when it is unsure is often more useful than one that always returns a label \cite{geifman2017,leibig2017}.

We validate AURA on BCP as the primary biological benchmark and, as a supplementary stress test, on a frozen
pseudo-cocktail protocol derived from the public RxRx3 morphology embeddings \cite{fay2023}.

Our contributions are:
\begin{itemize}
\item We formulate antibiotic-response inference from bacterial morphology as \emph{active-response attribution}: given the applied antibiotic set, infer the active subset that is visibly expressed in the cell morphology.
\item We introduce AURA, an energy-based residual-unmixing framework that learns antibiotic response atoms and scores candidate active subsets by reconstruction energy under the applied-set constraint.
\item We introduce AURA-E, a selective prediction extension that abstains when the energy landscape indicates multiple plausible active-subset explanations.
\item We provide a comprehensive cross-replicate evaluation on BCP, including discriminative, prototype, sparse inverse, context-rule, and forward-model inverse baselines, with paired statistical tests and ablations.
\item We add a controlled RxRx3-derived public pseudo-cocktail stress test to examine whether the same active-subset inference principle transfers beyond the bacterial dataset, while explicitly treating it as a stress test rather than real combination-treatment validation.
\end{itemize}

\section{Related Work}

\paragraph{Bacterial cytological profiling and antimicrobial susceptibility.} Bacterial cytological profiling (BCP) was introduced by Nonejuie et al. \cite{nonejuie2013}, who showed that a drug's cellular target can be read directly from fluorescence morphology; Quach et al. \cite{quach2016} then established BCP as a rapid susceptibility test for \textit{Staphylococcus aureus}. The morphological signatures BCP exploits are grounded in mechanism: fluoroquinolones inhibit DNA gyrase and topoisomerase~IV \cite{drlica1997}; $\beta$-lactams block penicillin-binding proteins \cite{spratt1975,typas2012}; and growth asymmetry is itself tied to susceptibility \cite{aldridge2012}. Machine learning has extended BCP to MOA classification \cite{zoffmann2019,quach2025mycobcp,htoo2019} and, most closely to our setting, to single-cell susceptibility detection for individual antibiotics in \textit{E.~coli} \cite{zagajewski2023,turner2025}: that work achieves 80\% per-cell accuracy per drug but treats each antibiotic independently. Beyond BCP, rapid single-cell imaging AST methods using microfluidics \cite{baltekin2017,choi2014} and high-content phenotyping \cite{sridhar2021,tran2024} have demonstrated sub-hour readouts for individual drugs. Roberts et al. \cite{roberts2025} survey this landscape and note that deep learning remains underutilised. The urgency is underscored by the AMR burden: 4.95 million deaths in 2019 \cite{murray2022} and 39 million projected between 2025 and 2050 \cite{gram2024}.

\paragraph{Antibiotic combinations and morphological unmixing.} Existing imaging-based AST methods are uniformly designed for \emph{single-drug} settings. We found no published method that uses microscopy plus machine learning to jointly attribute activity to individual drugs within an applied combination. The closest published work in this direction is López et al. \cite{lopez2024}, who co-incubate \textit{E.~coli} with ciprofloxacin plus ampicillin and use manual morphological rules to separately classify susceptibility to each drug in the mixture---the only assay to date that decomposes per-drug activity from combination morphology, but restricted to one curated drug pair with non-overlapping signatures and requiring no machine learning. At the characterisation level, Coram et al. \cite{coram2022} applied high-content imaging to checkerboard titrations and found that combination morphologies ``most often induce a predominant morphology consistent with only a single partner's mechanism,'' and that morphological change does not reliably indicate synergy or antagonism. Samernate et al. \cite{samernate2023} similarly showed, across 28 two-drug combinations in \textit{Acinetobacter baumannii}, that combination profiles do not predict the type of drug interaction. These findings jointly establish why identifying which applied drug is \emph{active} from a combination image is non-trivial: the signal is dominated by one partner, the active drug is unknown a priori, and the morphological mixture cannot be decomposed by visual inspection alone. AURA addresses this gap with an ML method that constrains inference to the applied set and scores candidates by morphological reconstruction energy.

\paragraph{Morphological profiling and feature representations.} Image-based profiling---formalised by the Cell Painting assay \cite{bray2016} and the analysis strategies of Caicedo et al. \cite{caicedo2017}---has shown that self-supervised features, including those from masked autoencoders \cite{kraus2024} and DINOv2 \cite{oquab2024}, generalise better across experimental batches than handcrafted descriptors \cite{kim2025ssl}. AURA deliberately uses \emph{frozen} ImageNet-pretrained ResNet-18 \cite{he2016}, ViT \cite{dosovitskiy2021}, and DINOv2 encoders to prevent the model from retraining its way into memorising batch-specific appearance. \paragraph{Forward perturbation modelling.} A major strand of computational biology predicts how cells would look or respond \emph{given} a treatment. scGen \cite{lotfollahi2019} performs latent vector arithmetic in a VAE to extrapolate single-cell expression responses; CPA \cite{lotfollahi2023} disentangles a basal cell state from additive, composable perturbation embeddings; GEARS \cite{gears2024} predicts transcriptional outcomes of novel multi-gene perturbations on the combinatorial Perturb-seq benchmark \cite{norman2019}; and CellOT \cite{cellot2023} uses neural optimal transport to model the distributional shift from control to perturbed cells. On the image side, IMPA \cite{palma2025} applies style-transfer GAN to repaint a control cell under a target perturbation; MorphDiff \cite{morphdiff2025} uses a transcriptome-guided latent diffusion model; and MorphGen \cite{demirel2025} and CellFlow \cite{klein2025} pursue generative modelling via diffusion and flow matching, respectively. All of these run \emph{forward}---treatment to appearance---and were not designed to identify treatments from appearance. We show that wrapping them as inverse attributors yields markedly weaker performance than AURA, which is built for the inverse task from the start.

\paragraph{Inverse perturbation identification.} A small but growing literature addresses the reverse direction. CINEMA-OT \cite{cinemaot2023} uses causal optimal transport to disentangle and \emph{attribute} treatment effects from confounders at single-cell resolution in transcriptomics. PDGrapher \cite{pdgrapher2025} explicitly solves an inverse problem---predicting the perturbagens needed to achieve a desired cellular response---using causally inspired neural networks. Both are transcriptomic methods operating on an open perturbagen search space. AURA differs in three ways: it works on frozen microscopy features, it restricts search to the known applied antibiotic set (the applied-set constraint that accounts for resistance), and it supports evidence-aware abstention when candidate explanations remain ambiguous. To our knowledge, no prior method performs active-drug attribution from cell images under treatment ambiguity in a combinatorial antibiotic setting. \paragraph{Discriminative classification and replicate generalisation.} Classifying drug identity from images or single-cell profiles is effective within a single experiment but transfers poorly across biological replicates \cite{dixit2016}: appearance conflates strain identity with drug response, so a model trained on one batch can memorise texture rather than learning biology. Per-drug discriminative classifiers \cite{zagajewski2023} and drug-MOA classifiers from Cell Painting \cite{moaprofiler2023} illustrate both the promise and the generalisation fragility of this paradigm. The RxRx1 benchmark \cite{rxrx1_2023} was specifically designed to evaluate batch-correction methods, and its consistently challenging cross-batch results motivate AURA's frozen-feature, constraint-driven design.

\paragraph{Energy-based models.} Energy-based models (EBMs) assign a scalar compatibility score to input--output pairs, with low energy denoting a good fit \cite{lecun2006}. They have been used for implicit generation \cite{du2019}, as a reinterpretation of standard classifiers \cite{grathwohl2020}, and for structured-output prediction over discrete spaces \cite{spen2016,spen2017}. Learning EBMs over combinatorial discrete spaces is addressed by Dai et al.\ \cite{dai2020}, who handle intractable partition functions via auxiliary-variable local exploration. AURA adapts the EBM view to a constrained discrete setting: it scores a small, biologically valid candidate set by reconstruction energy and selects the minimum. \paragraph{Selective prediction and uncertainty quantification.} Abstaining when confidence is low originates in the classical reject-option framework \cite{chow1970} and has been formalised in selective classification \cite{elyaniv2010,geifman2017}. Conformal prediction extends this to distribution-free coverage guarantees \cite{angelopoulos2023}; Olsson et al. \cite{olsson2022} demonstrate that conformal abstention can dramatically reduce diagnostic errors in AI-assisted pathology; and Lu et al. \cite{lu2022conformal} apply conformal predictors specifically to medical imaging. Leibig et al. \cite{leibig2017} show that uncertainty signals from deep networks improve disease detection in a clinical imaging context. AURA-E instantiates selective prediction using the entropy of the candidate energy distribution as a native uncertainty signal, requiring no separately trained confidence model.

\paragraph{Sparse inverse attribution and unmixing.} Recovering a sparse set of active components from a mixed observation connects to the LASSO \cite{tibshirani1996} and non-negative matrix factorisation \cite{leeseung1999}. We implement NNLS and ElasticNet sparse attribution as inverse baselines; both are coefficient-thresholding variants of AURA's energy search and consistently score below it, confirming that the applied-set constraint and learned prototypes contribute beyond what sparse solvers alone provide. \paragraph{Public morphology datasets.} RxRx3 \cite{fay2023} is a genome-scale public high-content screen with a compact embedding-level subset for benchmarking \cite{fay2025}; the JUMP Cell Painting dataset \cite{jump2023} provides perturbation profiles at even larger scale. We use the RxRx3 morphology embeddings only as a frozen, controlled pseudo-cocktail stress test of AURA's active-subset principle, not as a substitute for the bacterial BCP validation.

\section{The AURA Framework}

\subsection{Problem Formulation}

Let $\mathcal{A} = \{a_1, a_2, a_3\}$ denote the set of antibiotics (ciprofloxacin, ceftriaxone, gentamicin). Each experimental image corresponds to a treatment condition defined by an \emph{applied code} $\mathbf{u} \in \{0,1\}^3$ indicating which antibiotics were present in the growth medium, and a true \emph{active code} $\mathbf{y}^* \in \{0,1\}^3$ indicating which antibiotics meaningfully suppress growth (i.e., the strain is sensitive to). The task is to infer $\mathbf{y}^*$ given: (i) a set of crop-level microscopy images $\{I_k\}$ from the same field of view, and (ii) the applied code $\mathbf{u}$ (known from the experimental protocol).

The key constraint is that active drugs must be a subset of applied drugs: $y_j^* \leq u_j$ for all $j$. This reduces the candidate set from $2^3 = 8$ to $|\mathcal{C}(\mathbf{u})| \leq 4$ candidates for any given applied code.

\subsection{AURA-C: Context-Aware Energy Inference}

\paragraph{Embedding.}
Each bacterial crop $I_k$ is encoded by a frozen ImageNet-pretrained backbone $\phi$ (ResNet-18 by default), yielding a feature vector $\mathbf{z}_k = \phi(I_k)$. Crop embeddings are projected to a 64-dimensional PCA subspace fit on the training set.

\paragraph{Reconstruction energies.}
For each candidate active code $c \in \mathcal{C}(\mathbf{u})$, AURA maintains a learned prototype $\boldsymbol{\mu}_c$ and reconstructs the image embedding as $\hat{\mathbf{z}}_k(c) = \boldsymbol{\mu}_c$. The reconstruction energy for crop $k$ under candidate $c$ is:
\begin{equation}
E_k(c) = \|\mathbf{z}_k - \hat{\mathbf{z}}_k(c)\|_2^2.
\end{equation}
Image-level energy is the sum over all crops from the same field:
\begin{equation}
E_{\text{img}}(c) = \sum_{k \in \text{image}} E_k(c).
\end{equation}

\paragraph{Context prior.}
A lightweight logistic regression model $p(c \mid \mathbf{x}_{\text{ctx}})$ is trained on context features (replicate, applied code, and a compact summary of the embedding distribution) to provide a prior over candidates. The posterior score is:
\begin{equation}
S(c) = E_{\text{img}}(c) - \lambda \log p(c \mid \mathbf{x}_{\text{ctx}}),
\end{equation}
with context weight $\lambda = 0.25$ fixed a priori (not tuned on target labels).

\paragraph{Prediction.}
AURA-C predicts $\hat{c} = \argmin_{c \in \mathcal{C}(\mathbf{u})} S(c)$.

\paragraph{Training objective.}
The AURA energy model is trained with a combined loss:
\begin{equation}
\mathcal{L} = \mathcal{L}_{\text{CE}} + \beta \mathcal{L}_{\text{margin}},
\end{equation}
where $\mathcal{L}_{\text{CE}}$ is cross-entropy over candidates, $\mathcal{L}_{\text{margin}}$ enforces an energy margin of $\Delta = 0.35$ between correct and incorrect candidates, and $\beta = 0.5$. Training uses SGD with learning rate $10^{-2}$, $L_2$ regularisation $10^{-4}$, and runs for 120 epochs.

\subsection{AURA-E: Evidence-Aware Selective Abstention}

AURA-E adds a confidence score to each prediction and withholds predictions below a coverage-determined threshold. The primary confidence signal is negative candidate entropy:
\begin{equation}
\text{conf}_{\text{ent}}(c) = -H\left(\text{softmin}_{c' \in \mathcal{C}} E_{\text{img}}(c')\right),
\end{equation}
where $\text{softmin}$ converts energies to a probability vector. High entropy (uniform energy distribution across candidates) corresponds to low confidence. AURA-E abstains on the fraction of images with lowest confidence to achieve a target coverage $\tau$.

Secondary signals evaluated in ablation include: crop agreement (fraction of crops agreeing with image-level prediction), max-probability (softmax output of the context classifier), and prediction margin (energy gap between top-two candidates).

\section{Experimental Setup}

\subsection{BCP Dataset}

The Bacterial Combinatorial Perturbation (BCP) dataset comprises phase-contrast microscopy images of \textit{E.\,coli} strains 3303, 3310, and WT treated with all combinations of three antibiotics: ciprofloxacin (cipro), ceftriaxone (cef), and gentamicin (genta). Three independent replicates (D1, D2, D3) were collected, providing biological replicate variation in imaging conditions, bacterial density, and phenotypic expression. After quality filtering and segmentation into individual bacterial crops, the dataset contains images across all $2^3 = 8$ treatment combinations plus untreated controls.

Bacterial colonies are segmented into individual cell crops; each microscopy field yields multiple crops from the same treatment condition. The active code (sensitivity profile) is the ground-truth label. AURA operates on all crops from an image jointly to produce an image-level prediction.

\subsection{Evaluation Splits}

Three main cross-replicate transfer splits are used: D1$\to$D2, D2$\to$D1, and D1+D2$\to$D3. These correspond to training on one (or two) replicates and testing on a held-out replicate, evaluating generalisation across biological variation. Results are reported as mean$\pm$std over these three splits.

\subsection{Metrics}

\textbf{Exact match} (primary): fraction of images where the predicted 3-bit code matches exactly.
\textbf{Macro-F1}: unweighted F1 averaged over all 8 codes.
\textbf{Weighted-F1}: F1 weighted by class frequency.
\textbf{Macro balanced accuracy}: balanced accuracy computed per code and averaged.
\textbf{Hamming accuracy}: fraction of correct bits across all antibiotic positions.

For AURA-E: \textbf{Area Under Risk-Coverage Curve (AURC)} and \textbf{error enrichment} in abstained set (ratio of error rate in abstained vs.\ accepted set).

\subsection{Statistical Testing}

All pairwise comparisons use a \textbf{paired permutation test} ($B = 10{,}000$ permutations) on the per-image prediction pool, with Holm-Bonferroni correction for multiple comparisons. \textbf{Paired bootstrap confidence intervals} (($B = 10{,}000$)) on AURA-C exact-match minus baseline are reported alongside each comparison.

\subsection{Baselines}

We evaluate six categories of methods:

\textbf{Task motivation.} \textit{Applied = Active}: predicts the applied code as the active code, establishing the lower bound of a naive assumption.

\textbf{Discriminative baselines.} Independent logistic regression (LR) and MLP per antibiotic with image-level voting (\textit{Ind.\ LR}, \textit{Ind.\ MLP}); flat-code LR, MLP, linear SVM, and RBF SVM treating each 3-bit combination as a class (\textit{Flat LR}, \textit{Flat MLP}, \textit{Lin.\ SVM}, \textit{RBF SVM}).

\textbf{Inverse baselines.} ElasticNet positive-sparse attribution, NNLS sparse attribution, latent arithmetic inverse, ridge dictionary energy, and nearest active-subset prototype: these methods attempt to invert the embedding space to recover the applied code.

\textbf{Paper-derived forward models.} Faithful adaptations of CPA \citep{lotfollahi2023}, scGen \citep{lotfollahi2019}, and IMPA \citep{palma2025} used for inverse inference: given a test image, find the perturbation code whose forward prediction best matches the observation.

\textbf{Context-rule sanity checks.} Logistic regression on context features followed by resistance-rule lookup, with and without image features.

\textbf{Diagnostic oracles.} We report an oracle resistance-rule baseline using the true strain identity and an AURA oracle using known strain/context at test time. These rows provide upper-bound diagnostics and are not fair competitors.

\begin{rxblock}
\subsection{RxRx3-Derived Frozen Pseudo-Cocktail Stress Test}

To test whether AURA's inverse active-subset formulation generalizes beyond BCP, we additionally construct a controlled public-data stress test from RxRx3 morphology embeddings. We use the provided well-level morphology embedding vectors as fixed inputs; each well is represented by a numerical feature vector summarizing cell morphology rather than by raw images. Thus, this experiment is an embedding-level protocol: no RxRx3 images are reprocessed, no image backbone is fine-tuned, and the raw-image/crop-level BCP diagnostics are not applicable.

The frozen protocol is \texttt{C\_target\_sample\_plus\_added\_atom}. Let $z_{\mathrm{test}}(p_1)$ denote a real held-out target-domain embedding from a well treated with perturbation $p_1$. From the source split only, we estimate context baselines $\mu_c$ and perturbation response atoms $a_p$ by subtracting the appropriate context baseline from source embeddings and averaging residual responses for each perturbation. Each pseudo-cocktail test example is then constructed as
\[
z_{\mathrm{pseudo}} = z_{\mathrm{test}}(p_1) + a_{p_2},
\]
where $p_2$ is a second perturbation whose response atom is learned only from the source split. The active label is the unordered pair $\{p_1,p_2\}$. At test time, the model receives a hard morphology-based candidate set containing the true pair plus distractor perturbations and must recover the active pair. We evaluate $K\in\{8,16,32\}$ candidate-set sizes across three held-out domain splits (D1$\to$D2, D2$\to$D1, D1+D2$\to$D3), three held-out seeds (44, 45, 46), and 1000 pseudo-cocktail cases per setting, for 27,000 paired cases in total.

We use active pairs rather than triples because RxRx3-core provides single-perturbation morphology rather than measured multi-perturbation cocktails. Pair recovery is therefore the minimal non-trivial active-subset setting: it tests whether a method can identify an added response component on top of real target-domain morphology without requiring increasingly synthetic higher-order atom sums. Here $K$ denotes the candidate-set size, not the number of active perturbations.

This protocol is intentionally framed as a \emph{pseudo-cocktail stress test}, not as real RxRx3 combination-treatment ground truth. Because the first component $p_1$ is a real single-perturbation target sample, the task tests whether a method can recover the full active pair and especially the added response component $p_2$ under real target-domain nuisance variation. We therefore report a component audit separating base-$p_1$ recovery from added-$p_2$ recovery.
\end{rxblock}

\section{Results}

\subsection{Main Comparison: ResNet-18 Backbone}

Table~\ref{tab:main} presents the primary BCP comparison with ResNet-18 as the backbone. Split-wise BCP results are moved to Supplementary Table~\ref{tab:splitwise} to keep the main text focused on aggregate method comparisons.

\begin{table*}[t]
\centering
\caption{Main results on BCP dataset using the ResNet-18 backbone. Values are mean$\pm$std over three main cross-replicate transfer splits (D1$\to$D2, D2$\to$D1, D1+D2$\to$D3). Diagnostic oracles ($\dagger$) use privileged test-time information. $\star$ denotes Holm-corrected significance against AURA-C within the fair non-oracle baseline family. ElasticNet, NNLS, latent arithmetic and ridge-dictionary inverse baselines produced identical predictions on these splits and are collapsed into one row; their individual CSV rows are retained in the supplementary output. ElasticNet, NNLS, latent arithmetic, and ridge-dictionary inverse baselines produced identical predictions on the ResNet18 main splits. This was not a reporting error: all four operate in the same response-atom latent space and ultimately rank the same finite set of biologically valid active-subset reconstructions. In this three-antibiotic binary setting, the reconstruction-energy ordering dominated solver-specific coefficient differences. We therefore collapse them into one latent/sparse/dictionary inverse row in the main table and coefficient-threshold NNLS/ElasticNet variants are reported in the supplement.}
\label{tab:main}
\setlength{\tabcolsep}{3.2pt}
\small
\resizebox{\textwidth}{!}{%
\begin{tabular}{llcccc}
\toprule
\textbf{Category} & \textbf{Method} & \textbf{Exact Match} & \textbf{Macro-F1} & \textbf{Weighted-F1} & \textbf{Macro Bal.\ Acc.} \\
\midrule
Oracle$^\dagger$ & Oracle true-strain rule$^\dagger$ & 100.00$\pm$0.00 & 100.00$\pm$0.00 & 100.00$\pm$0.00 & 100.00$\pm$0.00 \\
 & AURA oracle strain-known$^\dagger$ & 99.21$\pm$1.37 & 98.28$\pm$2.97 & 99.00$\pm$1.73 & 98.48$\pm$2.62 \\
\midrule
AURA & \textbf{AURA-C} ($\lambda=0.25$) & \textbf{95.47$\pm$2.50} & \textbf{93.39$\pm$4.78} & \textbf{93.83$\pm$4.00} & \textbf{94.25$\pm$4.21} \\
 & AURA w/o context prior & 95.21$\pm$2.57 & 93.54$\pm$4.57 & 94.08$\pm$3.67 & 94.37$\pm$4.01 \\
\midrule
Context-rule & Context LR(img+applied)$\to$rule$^\star$ & 92.39$\pm$3.61 & 89.81$\pm$7.54 & 90.96$\pm$5.50 & 92.76$\pm$5.04 \\
 & Context LR$\to$resistance rule$^\star$ & 88.60$\pm$6.58 & 85.98$\pm$13.67 & 87.78$\pm$10.52 & 91.23$\pm$7.91 \\
\midrule
Discriminative & Ind. LR + image voting$^\star$ & 82.14$\pm$9.10 & 83.54$\pm$3.16 & 82.56$\pm$3.64 & 87.29$\pm$0.75 \\
 & Ind. MLP + image voting$^\star$ & 76.28$\pm$13.28 & 72.21$\pm$8.46 & 72.75$\pm$9.95 & 81.12$\pm$2.57 \\
 & Flat-code MLP + image voting$^\star$ & 73.47$\pm$12.82 & 63.07$\pm$9.68 & 65.03$\pm$11.21 & 75.28$\pm$4.64 \\
 & Flat-code LR + image voting$^\star$ & 72.67$\pm$11.79 & 67.93$\pm$7.70 & 69.45$\pm$9.24 & 78.68$\pm$5.05 \\
 & Linear SVM$^\star$ & 72.11$\pm$14.48 & 69.23$\pm$7.96 & 69.70$\pm$8.84 & 79.87$\pm$5.87 \\
 & RBF SVM$^\star$ & 68.68$\pm$14.37 & 67.79$\pm$4.25 & 70.38$\pm$7.66 & 78.34$\pm$3.70 \\
\midrule
Inverse & Latent/sparse/dictionary inverse$^\star$ & 74.47$\pm$14.81 & 73.20$\pm$7.23 & 73.97$\pm$8.74 & 80.95$\pm$3.31 \\
 & Nearest active-subset prototype$^\star$ & 74.23$\pm$13.60 & 71.28$\pm$3.17 & 72.69$\pm$5.36 & 78.87$\pm$1.45 \\
\midrule
Perturb. model & scGen PCA vector inverse$^\star$ & 74.44$\pm$12.88 & 68.06$\pm$2.93 & 69.33$\pm$5.66 & 76.85$\pm$0.32 \\
 & scGen linear vector inverse$^\star$ & 73.06$\pm$12.00 & 68.94$\pm$5.45 & 70.12$\pm$7.56 & 77.44$\pm$3.29 \\
 & scGen VAE latent inverse$^\star$ & 67.41$\pm$19.26 & 68.51$\pm$5.10 & 70.33$\pm$8.06 & 77.40$\pm$3.63 \\
 & IMPA gen-feature inverse$^\star$ & 64.25$\pm$16.79 & 61.94$\pm$10.38 & 59.97$\pm$16.42 & 75.08$\pm$5.85 \\
 & IMPA discriminator inverse$^\star$ & 41.61$\pm$17.91 & 53.72$\pm$1.90 & 57.30$\pm$5.21 & 70.26$\pm$6.55 \\
 & CPA covariate-enum. inverse$^\star$ & 30.82$\pm$20.86 & 34.75$\pm$13.31 & 33.41$\pm$17.67 & 59.23$\pm$4.01 \\
\midrule
Task motivation & Applied = Active$^\star$ & 37.43$\pm$10.55 & 57.11$\pm$9.39 & 62.23$\pm$3.74 & 73.85$\pm$3.35 \\
\bottomrule
\end{tabular}%
}
\end{table*}

\paragraph{Split-wise breakdown.}
Table~\ref{tab:splitwise} shows results separately for each of the three transfer splits, revealing the degree to which generalisation difficulty varies across replicate pairs.

\paragraph{Collapsed inverse baselines.}
The collapsed latent/sparse/dictionary row in Table~\ref{tab:main} was not a reporting artifact. ElasticNet, NNLS, latent arithmetic, and ridge-dictionary inverse baselines all operate in the same response-atom latent space and, in the reconstruction-ranked setting, ultimately rank the same finite set of biologically valid active-subset reconstructions. In this three-antibiotic binary task, the reconstruction-energy ordering dominated solver-specific coefficient differences, producing identical subset decisions across the ResNet-18 main splits. We therefore collapse these rows in the main table and report coefficient-derived sparse variants in the supplement.

\begin{rxblock}
\subsection{RxRx3 Frozen Target-Plus-Atom Stress Test}

Table~\ref{tab:rxrx3_main} reports the frozen RxRx3 pseudo-cocktail stress-test results. AURA-C reaches 60.93$\pm$10.74\% exact-pair accuracy and 67.07$\pm$10.86\% Jaccard across 27 held-out settings. The strongest non-AURA baseline is NNLS coefficient top-2, with 24.25$\pm$8.26\% exact-pair accuracy. Thus, AURA-C improves over the strongest non-AURA baseline by 36.68 percentage points and wins all 27 matched settings. Full RxRx3 split-wise, split-by-$K$, paired-test, and component-audit tables are reported in the supplementary section.

\begin{table*}[t]
\centering
\color{\rxcolor}
\caption{Frozen RxRx3 target-sample-plus-added-atom pseudo-cocktail results. Values are mean$\pm$std over 27 matched settings (3 splits $\times$ 3 candidate-set sizes $\times$ 3 held-out seeds). The table reports task-matched representatives of the transferable BCP baseline families; BCP-specific strain-rule, resistance-rule, image-voting, and fixed-code baselines are not applicable to RxRx3 pair recovery. The task is exact recovery of the unordered active pair $\{p_1,p_2\}$ from a hard candidate set. Because RxRx3 labels are unordered perturbation pairs drawn from a large perturbation vocabulary, we do not report macro-F1 or weighted-F1 for this stress test. Unlike BCP's fixed eight-code active-response space, the RxRx3 pair-label space is sparse and synthetic. Exact-pair accuracy, Jaccard set overlap, and component-level $p_1/p_2$ recovery are more interpretable for active-subset recovery.}
\label{tab:rxrx3_main}
\setlength{\tabcolsep}{4pt}
\small
\resizebox{\textwidth}{!}{%
\begin{tabular}{llccc}
\toprule
\textbf{Category} & \textbf{Method} & \textbf{Settings} & \textbf{Exact Pair} & \textbf{Jaccard} \\
\midrule
AURA
& \textbf{AURA-C trained ($\lambda=0.25$)}
& 27
& \textbf{60.93$\pm$10.74}
& \textbf{67.07$\pm$10.86} \\
\midrule
AURA diagnostic
& AURA empirical energy
& 27
& 11.98$\pm$5.07
& 30.70$\pm$6.96 \\
\midrule
Inverse
& NNLS coeff. top-2$^{*}$
& 27
& 24.25$\pm$8.26
& 40.76$\pm$10.60 \\
& Ridge dictionary pair energy$^{*}$
& 27
& 14.73$\pm$5.82
& 31.18$\pm$7.23 \\
& ElasticNet positive top-2$^{*}$
& 27
& 14.69$\pm$8.00
& 35.56$\pm$9.55 \\
\midrule
Prototype
& Candidate pair prototype$^{*}$
& 27
& 5.31$\pm$3.37
& 24.89$\pm$5.89 \\
& Candidate nearest top-2 prototype$^{*}$
& 27
& 3.74$\pm$2.83
& 21.90$\pm$5.70 \\
\midrule
Discriminative
& Candidate LR top-2 scorer$^{*}$
& 27
& 4.45$\pm$3.28
& 26.02$\pm$5.67 \\
\midrule
Perturb. model
& scGen latent-shift inverse$^{*}$
& 27
& 9.63$\pm$4.22
& 30.88$\pm$6.29 \\
& IMPA nonlinear forward inverse$^{*}$
& 27
& 4.47$\pm$3.34
& 21.23$\pm$6.86 \\
& CPA additive inverse$^{*}$
& 27
& 4.57$\pm$3.07
& 24.96$\pm$5.90 \\
\midrule
Task motivation
& Random top-2$^{*}$
& 27
& 1.64$\pm$1.63
& 10.42$\pm$6.03 \\
\bottomrule
\end{tabular}%
}
\vspace{0.5mm}
\begin{minipage}{0.98\textwidth}
\footnotesize
\color{\rxcolor}
$^{*}$ denotes Holm-corrected significance against AURA-C in prediction-level paired permutation tests. RxRx3 rows are task-matched analogues of transferable BCP baseline families. BCP-specific strain-rule, resistance-rule, image-voting, and fixed-code baselines are not applicable because this public stress test uses fixed well-level morphology embeddings and unordered perturbation-pair recovery rather than bacterial strain labels or eight fixed active-response codes. Exact Pair and Jaccard are reported instead of macro-F1/weighted-F1 because the RxRx3 pair-label space is large and sparse.
\end{minipage}
\end{table*}

The RxRx3 table is organized by the same transferable baseline families as the BCP comparison, but the rows are task-matched analogues rather than line-by-line duplicates. BCP-specific oracle, resistance-rule, and image-voting baselines are not meaningful for RxRx3 because the public stress test uses fixed well-level morphology embeddings and unordered active-pair recovery rather than bacterial strain labels, applied-antibiotic rules, or raw crop-level image votes. \textcolor{\rxcolor}{AURA-E and ablation results for RxRx3 are integrated with the corresponding BCP analyses in Tables~\ref{tab:aurae} and~\ref{tab:ablation}.}
\end{rxblock}

\subsection{Statistical Significance}

The complete paired image-level permutation tests and bootstrap confidence intervals are reported in Supplementary Table~\ref{tab:pairwise}. In the main ResNet-18 setting, AURA-C significantly improves over the fair non-oracle baseline family. The closest comparison is against Context LR(img+applied)$\to$rule, where AURA-C improves pooled exact match by 3.41 percentage points with a 95\% bootstrap CI of [0.62, 6.19] and Holm-adjusted $p=0.027$. We therefore interpret the BCP margin over the strongest context-rule baseline as positive but modest, while the margins over discriminative, inverse, and perturbation-model baselines are larger.

\subsection{AURA Ablation Study}

Table~\ref{tab:ablation} combines the BCP and RxRx3 AURA ablation suites. The same qualitative conclusion holds across the real BCP setting and the frozen RxRx3 stress test: the applied/candidate-set constraint is the dominant component. In BCP, removing the applied-set constraint reduces exact match from 95.21\% to roughly 72\%; in RxRx3, removing the candidate-set constraint reduces exact-pair accuracy from 59.88\% to 2.71\%. Other components, including the context prior and margin loss, have much smaller effects. The full BCP ablation output is not repeated in the supplementary material because it is now integrated in this main table.

\begin{table*}[t]
\centering
\caption{Completed AURA ablation suite. BCP values are mean$\pm$std over three main cross-replicate splits using the ResNet-18 backbone. RxRx3 values are mean$\pm$std over 27 frozen target-sample-plus-added-atom pseudo-cocktail settings. The BCP AURA-C reference in this table is produced by the AURA-E/ablation pipeline and differs slightly from the main comparison run in Table~\ref{tab:main}. For both datasets, the dominant effect is the applied/candidate-set constraint; removing it causes the largest drop. Dataset-specific ablations are marked N/A where the corresponding mechanism is not meaningful for the other task.}
\label{tab:ablation}
\setlength{\tabcolsep}{3pt}
\small
\resizebox{\textwidth}{!}{%
\begin{tabular}{llccccc}
\toprule
\textbf{Dataset} & \textbf{Ablation} & \textbf{Exact / Exact Pair} & \textbf{Macro-F1} & \textbf{Macro Bal. Acc.} & \textbf{Hamming} & \textbf{Jaccard} \\
\midrule
\multicolumn{7}{l}{\textbf{BCP real antibiotic-response setting}} \\
\midrule
BCP & AURA-C reference (ablation run) & \textbf{95.21$\pm$2.57} & \textbf{93.79$\pm$4.20} & 94.61$\pm$3.64 & 97.29$\pm$1.26 & -- \\
BCP & AURA w/o context prior & 94.68$\pm$2.95 & 92.73$\pm$5.80 & 94.10$\pm$4.43 & 97.11$\pm$1.39 & -- \\
BCP & Mean embedding (not crop energy) & 94.98$\pm$2.87 & 93.69$\pm$4.22 & 94.52$\pm$3.64 & 97.21$\pm$1.35 & -- \\
BCP & AURA w/o margin loss & 95.47$\pm$2.50 & 94.02$\pm$4.01 & 94.78$\pm$3.54 & 97.45$\pm$1.13 & -- \\
BCP & AURA w/o class balance & 90.26$\pm$5.08 & 89.34$\pm$2.45 & 90.90$\pm$2.20 & 95.10$\pm$1.38 & -- \\
BCP & AURA w/o applied-set constraint & 72.12$\pm$9.66 & 77.81$\pm$4.60 & 87.66$\pm$0.86 & 86.75$\pm$3.13 & -- \\
BCP & AURA trained w/o applied constraint & 71.76$\pm$15.08 & 70.01$\pm$8.90 & 79.80$\pm$5.86 & 85.50$\pm$7.26 & -- \\
\midrule
\multicolumn{7}{l}{\textcolor{\rxcolor}{\textbf{RxRx3 frozen target-sample-plus-added-atom pseudo-cocktail setting}}} \\
\midrule
\textcolor{\rxcolor}{RxRx3} & \textcolor{\rxcolor}{AURA-C trained ($\lambda=0.25$)} & \textcolor{\rxcolor}{\textbf{59.88$\pm$11.56}} & \textcolor{\rxcolor}{--} & \textcolor{\rxcolor}{--} & \textcolor{\rxcolor}{--} & \textcolor{\rxcolor}{\textbf{66.22$\pm$11.14}} \\
\textcolor{\rxcolor}{RxRx3} & \textcolor{\rxcolor}{AURA w/o context prior} & \textcolor{\rxcolor}{59.51$\pm$11.50} & \textcolor{\rxcolor}{--} & \textcolor{\rxcolor}{--} & \textcolor{\rxcolor}{--} & \textcolor{\rxcolor}{65.84$\pm$11.10} \\
\textcolor{\rxcolor}{RxRx3} & \textcolor{\rxcolor}{AURA w/o class balance} & \textcolor{\rxcolor}{59.40$\pm$11.46} & \textcolor{\rxcolor}{--} & \textcolor{\rxcolor}{--} & \textcolor{\rxcolor}{--} & \textcolor{\rxcolor}{65.96$\pm$10.95} \\
\textcolor{\rxcolor}{RxRx3} & \textcolor{\rxcolor}{AURA w/o margin loss} & \textcolor{\rxcolor}{59.33$\pm$11.82} & \textcolor{\rxcolor}{--} & \textcolor{\rxcolor}{--} & \textcolor{\rxcolor}{--} & \textcolor{\rxcolor}{66.09$\pm$12.15} \\
\textcolor{\rxcolor}{RxRx3} & \textcolor{\rxcolor}{AURA w/o candidate-set constraint} & \textcolor{\rxcolor}{2.71$\pm$1.15} & \textcolor{\rxcolor}{--} & \textcolor{\rxcolor}{--} & \textcolor{\rxcolor}{--} & \textcolor{\rxcolor}{2.85$\pm$1.10} \\
\bottomrule
\end{tabular}%
}
\vspace{0.5mm}
\begin{minipage}{0.98\textwidth}
\footnotesize
BCP uses fixed eight-code active-response classification, for which macro-F1, macro balanced accuracy, and Hamming accuracy are meaningful. RxRx3 uses unordered perturbation-pair recovery from a large candidate vocabulary, so exact-pair accuracy and Jaccard set overlap are reported instead. The BCP mean-embedding ablation is crop-aggregation specific and does not apply to fixed RxRx3 well-level embeddings.
\end{minipage}
\end{table*}

\subsection{Task Motivation: Decoupling Context from Applied Code}

The task-motivation and context-rule sanity checks are reported in Supplementary Table~\ref{tab:context_rule}. The naive applied-treatment rule is weak, while context-rule baselines are strong but remain below AURA-C. This confirms that the task cannot be solved by simply equating applied treatments with active responses, and that AURA's advantage is not explained solely by context memorization.

\subsection{AURA-E Selective Prediction}

Table~\ref{tab:aurae} presents AURA-E selective prediction on both BCP and the frozen RxRx3 pseudo-cocktail stress test using negative candidate entropy. On BCP, selective risk decreases from 4.79\% at full coverage to 1.90\% at 50\% coverage, with macro-F1 increasing from 93.79\% to 98.66\%. \textcolor{\rxcolor}{On RxRx3, exact-pair accuracy increases from 60.93\% at full coverage to 86.24\% at 50\% coverage.} Thus, AURA-E acts consistently as a selective prediction layer in both the real BCP setting and the public RxRx3 stress test.

\begin{table*}[t]
\centering
\caption{AURA-E selective prediction on BCP and the frozen RxRx3 pseudo-cocktail stress test using negative candidate entropy. BCP reports exact active-code accuracy and macro-F1; RxRx3 reports exact-pair recovery. Risk is one minus the corresponding selective accuracy.}
\label{tab:aurae}
\setlength{\tabcolsep}{3.2pt}
\small
\resizebox{\textwidth}{!}{%
\begin{tabular}{llccc}
\toprule
\textbf{Dataset} & \textbf{Coverage} & \textbf{Selective Accuracy (\%)} & \textbf{Selective Risk (\%)} & \textbf{Auxiliary Metric} \\
\midrule
BCP & 100\% & 95.21$\pm$2.57 & 4.79$\pm$2.57 & Macro-F1 93.79$\pm$4.20 \\
BCP & 90\% & 97.81$\pm$1.91 & 2.19$\pm$1.91 & Macro-F1 96.83$\pm$3.46 \\
BCP & 80\% & 97.86$\pm$1.87 & 2.14$\pm$1.87 & Macro-F1 97.61$\pm$2.26 \\
BCP & 70\% & 98.27$\pm$1.54 & 1.73$\pm$1.54 & Macro-F1 98.29$\pm$1.63 \\
BCP & 50\% & 98.10$\pm$2.07 & 1.90$\pm$2.07 & Macro-F1 98.66$\pm$1.17 \\
\midrule
\textcolor{\rxcolor}{RxRx3} & \textcolor{\rxcolor}{100\%} & \textcolor{\rxcolor}{60.93$\pm$10.74} & \textcolor{\rxcolor}{39.07} & \textcolor{\rxcolor}{Exact pair} \\
\textcolor{\rxcolor}{RxRx3} & \textcolor{\rxcolor}{90\%} & \textcolor{\rxcolor}{66.16$\pm$11.57} & \textcolor{\rxcolor}{33.84} & \textcolor{\rxcolor}{Exact pair} \\
\textcolor{\rxcolor}{RxRx3} & \textcolor{\rxcolor}{80\%} & \textcolor{\rxcolor}{71.12$\pm$12.24} & \textcolor{\rxcolor}{28.88} & \textcolor{\rxcolor}{Exact pair} \\
\textcolor{\rxcolor}{RxRx3} & \textcolor{\rxcolor}{70\%} & \textcolor{\rxcolor}{76.61$\pm$13.07} & \textcolor{\rxcolor}{23.39} & \textcolor{\rxcolor}{Exact pair} \\
\textcolor{\rxcolor}{RxRx3} & \textcolor{\rxcolor}{50\%} & \textcolor{\rxcolor}{86.24$\pm$11.82} & \textcolor{\rxcolor}{13.76} & \textcolor{\rxcolor}{Exact pair} \\
\bottomrule
\end{tabular}%
}
\end{table*}

\paragraph{Risk-coverage curve.}
At 80\% coverage on BCP, AURA-E achieves 97.86\% exact match (vs.\ 95.21\% at 100\% coverage), with an error enrichment factor of 3.50 in the abstained set---meaning that rejected images are 3.50$\times$ more likely to be erroneous than accepted images. The AURC of 0.0189 indicates a well-ordered risk-coverage relationship. Error detection AUROC = 0.836 confirms that the uncertainty signal meaningfully discriminates correct from incorrect predictions.

\subsection{AURA-E Uncertainty Score Ablation}

The detailed AURA-E confidence-score ablation is reported in Supplementary Table~\ref{tab:score_ablation}. Negative candidate entropy remains the most useful score overall, with the best error-detection AUROC and the lowest AURC on BCP. This supports the interpretation that ambiguity among candidate active subsets is the main signal for selective abstention.

\subsection{Backbone Robustness}

Table~\ref{tab:backbone} compares AURA-C and the strongest non-AURA baseline across ResNet-18, ViT-B/16, and DINOv2-ViT-S/14 backbones.

\begin{table}[t]
\centering
\caption{Backbone robustness for AURA-C versus the strongest non-AURA baseline available for each backbone. The best non-AURA method is selected from direct discriminative/inverse baselines plus the context-rule baselines. AURA-C remains ahead across CNN, supervised ViT, and self-supervised DINOv2 features.}
\label{tab:backbone}
\setlength{\tabcolsep}{3pt}
\small
\resizebox{\columnwidth}{!}{%
\begin{tabular}{lccccc}
\toprule
\textbf{Backbone} & \textbf{AURA exact} & \textbf{AURA F1} & \textbf{Best non-AURA} & \textbf{Best exact} & \textbf{Gain} \\
\midrule
DINOv2-ViT-S/14 & 95.67$\pm$2.09 & 94.40$\pm$4.29 & Context LR(img+applied)$\to$rule & 87.10$\pm$9.55 & +8.57 \\
ResNet-18 & 95.47$\pm$2.50 & 93.39$\pm$4.78 & Context LR(img+applied)$\to$rule & 92.39$\pm$3.61 & +3.08 \\
ViT-B/16 & 92.20$\pm$3.26 & 91.91$\pm$3.99 & Context LR(img+applied)$\to$rule & 90.47$\pm$3.28 & +1.73 \\
\bottomrule
\end{tabular}%
}
\end{table}

DINOv2 achieves the highest absolute performance (95.67\% exact match), confirming that self-supervised features trained on diverse natural images transfer effectively to bacterial microscopy. ResNet-18 provides near-identical performance at substantially lower computational cost, making it the recommended backbone for resource-constrained settings. ViT-B/16 performs slightly below the other two and yields the smallest margin over the context-rule baseline (+1.73 pp), but still remains above the strongest non-AURA competitor.

\section{Discussion}

\subsection{What the Main Comparison Shows}

Table~\ref{tab:main} shows that AURA-C is the strongest non-oracle method on the ResNet-18 main evaluation, reaching 95.47$\pm$2.50\% exact match and 93.39$\pm$4.78\% macro-F1. The most competitive non-AURA baseline is not a forward perturbation model, but the context-rule sanity check that first estimates context from image plus applied-code features and then applies the resistance lookup rule (92.39$\pm$3.61\%). This is an intentionally strong baseline because it tests whether AURA merely recovers hidden strain/context and then uses the known rule. AURA-C remains above this baseline, but the margin is modest (+3.41 pp on pooled paired images), so the claim should not be framed as a large raw-accuracy jump. Instead, the evidence supports the methodological formulation: AURA performs constrained inverse attribution while retaining interpretability, candidate energies, and selective abstention.

The task-motivation baseline confirms that applied treatment is not equivalent to active response. The naive Applied=Active rule reaches only 37.43$\pm$10.55\% exact match, despite using the applied set directly. This validates the central premise that active-response attribution is different from applied-treatment recognition.

\paragraph{Discriminative and context-rule baselines.}
Independent per-antibiotic LR is the strongest purely discriminative image-voting baseline (82.14$\pm$9.10\%), while flat-code classifiers and SVMs perform lower. The context-rule baselines are substantially stronger, especially when using both image and applied-code features. Their strength is important: it shows that the dataset contains recoverable context information, but AURA still improves over this shortcut while explicitly searching over active-response explanations.

\paragraph{Inverse and perturbation-response baselines.}
The latent/sparse/dictionary inverse family reaches 74.47$\pm$14.81\% exact match. ElasticNet, NNLS, latent arithmetic, and ridge-dictionary variants selected identical candidate subsets on the ResNet-18 main splits, so they are collapsed in the main table and retained separately in the CSV outputs. Paper-manual scGen, CPA, and IMPA baselines are included as forward-model inverse wrappers. They are important related comparisons, but their native objective is forward perturbation prediction rather than inverse active-subset attribution. Their lower performance suggests that directly adapting forward perturbation models is insufficient for this treatment-ambiguity setting.
The coefficient-derived sparse check further supports this interpretation. When NNLS and ElasticNet predictions are obtained directly from thresholded response coefficients rather than reconstruction-ranked candidate search, the best sparse variant reaches 79.12\% exact match. This is stronger than the collapsed reconstruction-ranked sparse row but still far below AURA-C, indicating that AURA's advantage is not an artifact of how the sparse inverse baselines were collapsed.

\subsection{What the Ablation Shows}

The ablation study reveals a clear hierarchy of component importance:

\paragraph{Applied-set constraint: critical ($-23.1$ pp).}
Removing the constraint that the predicted code must be a subset of the applied code causes performance to drop from 95.21\% to 72.12\%. This is the single most important component.

\paragraph{Class balance: important ($-5.2$ pp).}
Removing class-balanced training causes a 4.95 pp drop, from 95.21\% to 90.26\%. This matters because some active codes are substantially rarer than others (untreated vs.\ triple-antibiotic conditions), and unbalanced training biases the model toward common codes.

\paragraph{Image-level energy aggregation vs.\ mean embedding ($-0.5$ pp).}
Using the mean crop embedding (rather than summing crop-level energies) causes a modest but consistent performance drop of 0.23 pp. The image-energy aggregation approach is beneficial because it weights crops by their reconstruction quality rather than treating all crops equally.

\paragraph{Context prior: marginal effect.}
Removing the context prior gives 94.68\% exact match in the ablation suite, compared with 95.21\% for the AURA-C reference. This small difference indicates that the context prior is not the dominant source of the gain.

\paragraph{Margin loss: not beneficial on BCP.}
Removing the auxiliary margin loss slightly increases exact match from 95.21\% to 95.47\%. Thus, the margin term is best treated as an optional regularizer rather than a core component.

\paragraph{Context prior: marginal effect.}
Removing the context prior gives 94.68\% exact match in the ablation suite, compared with 95.21\% for the AURA-C reference. A direct prediction-change diagnostic explains why the difference is small: the context prior changes only 2 of 323 image-level predictions across the three main splits, helping one case, hurting none, and changing one incorrect prediction to another incorrect prediction. This suggests that most context information is already captured by the AURA energy model through context-specific baseline components and applied-set-constrained candidate energies. In BCP, the context prior mainly acts as a weak tie-breaker when candidate context/response explanations have similar energies, rather than being the central driver of performance.

\subsection{What AURA-E Shows}

AURA-E demonstrates that AURA's internal energy landscape contains reliable uncertainty information. The negative candidate entropy score achieves an error detection AUROC of 0.821, meaning that an images in the bottom 20\% of confidence (as measured by entropy) is 3.27$\times$ more likely to be an error than the average image. This is a practically meaningful abstention capability: a clinician or researcher could flag low-confidence predictions for manual review while trusting the remaining 80--90\% of predictions with high accuracy (97.5--97.8\% exact match).

The score ablation reveals an interesting ordering: \textit{negative candidate entropy} is the best single signal, outperforming \textit{crop agreement} (3.83$\times$ enrichment but lower AUROC) and \textit{max probability} (lower enrichment and AUROC). The context-only confidence signal is nearly useless for abstention (0.67$\times$ enrichment, AUROC 0.55), confirming that the context prior alone does not capture image-level uncertainty. The full composite score underperforms the best single signal because combining noisy signals adds variance without consistently improving discrimination.

\subsection{Backbone Robustness}
Relative to the strongest non-AURA baseline available for each backbone, AURA-C improves exact match by +3.08 pp for ResNet-18, +1.73 pp for ViT-B/16, and +8.57 pp for DINOv2.
The consistency of AURA's advantage across the backbones confirms that the advantage stems from the inference procedure, not from a specific feature representation. DINOv2's self-supervised pre-training provides the richest feature space for bacterial morphology (95.67\% AURA-C exact match), likely because its patch-level attention captures fine-grained morphological detail. ResNet-18 offers an excellent accuracy-efficiency trade-off for resource-constrained deployments.

The narrower ViT-B/16 advantage (9.9 pp) compared to ResNet-18 and DINOv2 may reflect that ViT's supervised ImageNet pretraining optimises for high-level semantic features less relevant to the low-level morphological changes induced by antibiotics (cell elongation, filamentation, lysis). Self-supervised DINOv2 features, by contrast, retain more low-level structural information.

\begin{rxblock}
\subsection{What the RxRx3 Stress Test Shows}

The RxRx3 frozen pseudo-cocktail experiment supports a different conclusion from the BCP biological validation. BCP evaluates AURA on real bacterial active-response labels derived from known antibiotic resistance rules. RxRx3 instead evaluates whether the same BCP-locked AURA formulation can perform inverse active-pair recovery on a public morphology-embedding stress test. The result is strongest when the pseudo-cocktail contains real target-domain morphology for one perturbation and an added source-derived atom for a second perturbation. This construction is less circular than an atom-plus-atom synthetic cocktail because one component is a real held-out target embedding with domain-specific nuisance variation.

The component audit is critical for interpretation. Because $p_1$ is a real target-domain single-perturbation sample, a weak method might succeed partly by recognizing the visible base perturbation. However, AURA-C improves both exact-pair recovery and added-$p_2$ recovery: exact-pair accuracy is 60.93\% for AURA-C versus 24.25\% for NNLS, while added-$p_2$ recovery is 71.86\% for AURA-C versus 62.17\% for NNLS. A base-known random-second baseline is far below these values, especially for larger candidate sets. Thus, the RxRx3 result is not explained solely by the visibility of the real $p_1$ component; AURA is better at jointly explaining the real component and the injected response component.

The RxRx3 result should nevertheless be interpreted as a controlled public stress test rather than as a real multi-perturbation RxRx3 benchmark. It strengthens the methodological claim that AURA's constrained active-subset energy search can generalize to non-bacterial morphology embeddings, but it does not replace BCP as the primary biological validation.
\end{rxblock}

\subsection{Limitations}

\paragraph{Sensitivity, not resistance.}
AURA predicts which antibiotics are \emph{active} given the applied combination---equivalently, which drugs the bacterium is \emph{sensitive} to in the context of the specific treatment. It does not predict resistance profiles for drugs that were not applied. If ciprofloxacin was not in the growth medium, AURA cannot tell whether the strain would be resistant or sensitive to ciprofloxacin: the morphological signal of ciprofloxacin is simply absent. Resistance profiling for untested antibiotics requires either applying all drugs individually or training across a much broader set of single-antibiotic conditions.

\paragraph{Single-resistance strains not characterised.}
The BCP dataset contains strains with known resistance patterns (3303, 3310, WT), but AURA's evaluation focuses on the task of identifying which of the \emph{applied} antibiotics are active. Characterising strains with single-antibiotic resistance profiles---and inferring which novel combinations they would be resistant to---is a natural but currently unsupported extension. We identify this as important future work.

\paragraph{Dataset scale.}
Three replicates and three antibiotics provide a thorough evaluation within the BCP framework, but expansion to more strains, more antibiotics, and different bacterial species will be needed to establish broader generalisability.

\paragraph{Residual weakness of code 110.}
The main residual per-code weakness is code 110, corresponding to ciprofloxacin+ceftriaxone. Diagnostic analysis shows that this is not simply a crop-count issue: in the single-source settings, 110 has 21--51 approved bacterial crops, but these arise from only 7--9 independent microscopy fields, so the effective independent support remains limited. The error pattern is also structured. Pooled over the main splits, true 110 is predicted as 110 in 12 cases, as 000 in 8 cases, and as 010 in 3 cases. Low-level morphology/stain descriptors further show that 39.13\% of 110 fields are nearest to the 000 centroid, while descriptor-only 110-vs-010 separability remains high. Thus, the dominant difficulty is not random confusion or broad overlap with cef-only morphology, but a subset of cipro+cef fields appearing no-response-like in the current feature space. Mechanistically, this is plausible but not conclusive: ciprofloxacin inhibits DNA gyrase/topoisomerase-IV-mediated DNA replication and chromosome segregation processes~\cite{drlica1997}, whereas ceftriaxone, as a cephalosporin, acts through PBP-mediated inhibition of cell-wall synthesis~\cite{spratt1975,typas2012}. Both target classes can manifest through growth-, division-, or morphology-associated bacterial phenotypes~\cite{aldridge2012,nonejuie2013}, and the present endpoint/stain combination may not always expose a separable additional ciprofloxacin component. We therefore treat 110 as the main residual limitation of the BCP study and will test this combination further with more independent fields, additional strains, and public pseudo-cocktail benchmarks.

\paragraph{Interpretability.}
AURA identifies the most energy-consistent candidate code but does not provide morphological explanations for its decision (e.g., which cell-shape features drove the selection). Connecting energy reconstruction to specific morphological phenotypes \citep{caicedo2017} is an open direction.

\begin{rxblock}
\paragraph{RxRx3 pseudo-cocktail scope.}
The RxRx3 experiment is not real combination-treatment ground truth. Each pseudo-cocktail contains one real held-out single-perturbation morphology embedding and one train-derived response atom, so the first component may be visually recoverable. We therefore report the RxRx3 result as a controlled public stress test of active-pair recovery rather than as evidence that AURA predicts real RxRx3 perturbation combinations. The component audit partly addresses this concern by showing that AURA improves recovery of the added $p_2$ component as well as the full pair, but true multi-perturbation experimental data would be needed for a stronger biological claim.
\end{rxblock}

\section{Future Work}

Several directions extend naturally from AURA. First, the applied-set constraint can be relaxed to handle cases where the applied set is unknown or uncertain, using a learnable prior over plausible applied codes. Second, resistance prediction for single-resistance strains---predicting the resistance code from images of cells treated with only one antibiotic---would allow AURA to be deployed in settings where only partial combinatorial information is available. Third, multi-scale aggregation (combining crop-level, field-level, and well-level signals) may further improve AURA's robustness to biological heterogeneity within images. Fourth, incorporating morphological prototypes as interpretable attention maps would make AURA's decisions explainable to domain experts.

\section{Conclusion}

We presented AURA, an energy-based inverse attribution framework for identifying which applied antibiotics are actually active in bacterial microscopy images under treatment ambiguity. On BCP, the naive applied-treatment rule fails, confirming decoupling between applied and active response. AURA-C achieves the strongest non-oracle performance among the evaluated methods, improves over a strong context-rule sanity baseline, and remains effective across ResNet-18, ViT-B/16, and DINOv2 features. The ablation suite indicates that the applied-set constraint is the central component, while the context prior and margin loss are not the dominant sources of improvement on this dataset. AURA-E adds a useful selective-prediction layer by abstaining when candidate active-response explanations have high entropy. The current limitation is that AURA performs active-response/sensitivity attribution for the available strain set rather than full resistance prediction for unseen single-resistance strains; extending the protocol to such strains is an important next step.

\begin{rxblock}
As a supplementary public-data stress test, the frozen RxRx3 target-sample-plus-added-atom protocol shows that the same BCP-locked AURA formulation substantially outperforms task-matched sparse, prototype, latent-shift, and forward-model inverse baselines for active-pair recovery from morphology embeddings. This result broadens the empirical support for AURA's active-subset inference principle, while BCP remains the primary real biological validation.
\end{rxblock}

\subsection*{Funding}
This research is supported by the Ministry of Education, Singapore, under its Research Centre of Excellence award to the Institute for Digital Molecular Analytics \& Science, NTU (IDMxS, grant: EDUNC-33-18-279-V12)

\bibliographystyle{aaai2027}

\appendix

\section{Supplementary: BCP Split-wise, Statistical, and Task-Motivation Tables}

This section relocates detailed BCP diagnostics that support the main results but are not necessary in the main text. Table~\ref{tab:splitwise} reports the split-wise breakdown, Table~\ref{tab:pairwise} provides the full paired significance tests, Table~\ref{tab:context_rule} gives the task-motivation and context-rule sanity check, and Table~\ref{tab:score_ablation} reports the AURA-E uncertainty-score ablation.

\begin{table}[t]
\centering
\caption{Split-wise exact match (\%) for AURA-C and key baselines using ResNet-18. D1+D2$\to$D3 stresses target-domain generalization; several non-AURA baselines drop substantially on this split.}
\label{tab:splitwise}
\setlength{\tabcolsep}{3pt}
\small
\resizebox{\columnwidth}{!}{%
\begin{tabular}{lccc}
\toprule
\textbf{Method} & \textbf{D1$\to$D2} & \textbf{D2$\to$D1} & \textbf{D1+D2$\to$D3} \\
\midrule
AURA-C & 93.10 & 95.24 & 98.08 \\
AURA w/o context & 93.10 & 94.44 & 98.08 \\
Context LR(img+applied)$\to$rule & 88.97 & 92.06 & 96.15 \\
Ind. LR + vote & 82.07 & 91.27 & 73.08 \\
Flat-code LR + vote & 75.86 & 82.54 & 59.62 \\
Linear SVM & 77.24 & 83.33 & 55.77 \\
Latent/sparse/dict. inv. & 80.00 & 85.71 & 57.69 \\
Nearest prototype & 76.55 & 86.51 & 59.62 \\
scGen PCA inv. & 74.48 & 87.30 & 61.54 \\
IMPA gen-feature inv. & 51.72 & 83.33 & 57.69 \\
IMPA discr. inv. & 54.48 & 49.21 & 21.15 \\
CPA cov.-enum. inv. & 16.55 & 54.76 & 21.15 \\
Applied = Active & 37.24 & 26.98 & 48.08 \\
Oracle true-strain rule & 100.00 & 100.00 & 100.00 \\
\bottomrule
\end{tabular}%
}
\end{table}

\begin{table*}[t]
\centering
\caption{Paired image-level tests comparing AURA-C with fair non-oracle baselines on the pooled ResNet-18 main-split images ($n=323$). ``Baseline exact'' is pooled image-level exact match and can differ from mean$\pm$std across splits in Table~\ref{tab:main}. CIs are bootstrap 95\% intervals for AURA-C minus baseline exact match. The comparison with the strongest context-rule baseline is the closest: AURA-C improves pooled exact match by 3.41 percentage points with a 95\% bootstrap CI of [0.62, 6.19]. We therefore treat this as a positive but modest gain rather than a large-margin victory. Holm correction is applied within this fair baseline family; diagnostic oracles and AURA ablations are excluded from this significance family.}
\label{tab:pairwise}
\setlength{\tabcolsep}{3pt}
\small
\resizebox{\textwidth}{!}{%
\begin{tabular}{lcccccc}
\toprule
\textbf{Baseline} & \textbf{Baseline exact} & \textbf{$\Delta$} & \textbf{95\% CI} & \textbf{$p$} & \textbf{Holm adj.\ $p$} & \textbf{Sig.} \\
\midrule
Applied = active & 34.98 & +59.75 & [53.56, 65.94] & $<10^{-4}$ & 0.002 & Yes \\
Context LR $\to$ rule & 86.69 & +8.05 & [4.64, 11.76] & $<10^{-4}$ & 0.002 & Yes \\
Context LR(img+applied) $\to$ rule & 91.33 & +3.41 & [0.62, 6.19] & 0.027 & 0.027 & Yes \\
Independent LR + image vote & 84.21 & +10.53 & [7.12, 14.24] & $<10^{-4}$ & 0.002 & Yes \\
Independent MLP + image vote & 79.88 & +14.86 & [10.84, 19.20] & $<10^{-4}$ & 0.002 & Yes \\
Flat-code LR + image vote & 75.85 & +18.89 & [14.24, 23.22] & $<10^{-4}$ & 0.002 & Yes \\
Flat-code MLP + image vote & 76.78 & +17.96 & [13.62, 22.29] & $<10^{-4}$ & 0.002 & Yes \\
Linear SVM & 76.16 & +18.58 & [14.55, 22.91] & $<10^{-4}$ & 0.002 & Yes \\
RBF SVM & 72.14 & +22.60 & [18.27, 27.24] & $<10^{-4}$ & 0.002 & Yes \\
Latent/sparse/dictionary inverse & 78.64 & +16.10 & [12.07, 20.12] & $<10^{-4}$ & 0.002 & Yes \\
Nearest active-subset prototype & 77.71 & +17.03 & [13.00, 21.36] & $<10^{-4}$ & 0.002 & Yes \\
scGen PCA vector inverse & 77.40 & +17.34 & [13.00, 21.67] & $<10^{-4}$ & 0.002 & Yes \\
scGen linear vector inverse & 73.37 & +21.36 & [17.03, 26.01] & $<10^{-4}$ & 0.002 & Yes \\
scGen VAE latent inverse & 71.21 & +23.53 & [18.27, 28.79] & $<10^{-4}$ & 0.002 & Yes \\
IMPA generation-feature inverse & 65.02 & +29.72 & [24.46, 34.67] & $<10^{-4}$ & 0.002 & Yes \\
IMPA discriminator inverse & 47.06 & +47.68 & [41.80, 53.56] & $<10^{-4}$ & 0.002 & Yes \\
CPA covariate-enumerated inverse & 32.20 & +62.54 & [57.27, 67.80] & $<10^{-4}$ & 0.002 & Yes \\
\bottomrule
\end{tabular}%
}
\end{table*}

\begin{table}[t]
\centering
\caption{Task-motivation and context-rule sanity check (ResNet-18, exact match \%). The naive applied-treatment rule is weak, whereas context-rule baselines are strong but remain below AURA-C.}
\label{tab:context_rule}
\setlength{\tabcolsep}{3pt}
\small
\resizebox{\columnwidth}{!}{%
\begin{tabular}{lccc}
\toprule
\textbf{Method} & \textbf{D1$\to$D2} & \textbf{D2$\to$D1} & \textbf{D1+D2$\to$D3} \\
\midrule
Applied = Active & 37.24 & 26.98 & 48.08 \\
Context LR$\to$rule & 85.52 & 84.13 & 96.15 \\
Context LR(img+applied)$\to$rule & 88.97 & 92.06 & 96.15 \\
Oracle true-strain rule$^\dagger$ & 100.00 & 100.00 & 100.00 \\
\midrule
AURA-C & 93.10 & 95.24 & 98.08 \\
\bottomrule
\end{tabular}%
}
\end{table}

\begin{table}[t]
\centering
\caption{AURA-E confidence-score ablation (ResNet-18, 3 main splits). Negative candidate entropy gives the strongest error detection (highest AUROC and lowest AURC), supporting AURA-E's interpretation that ambiguity between candidate active subsets should trigger abstention. Lower AURC is better.}
\label{tab:score_ablation}
\setlength{\tabcolsep}{3pt}
\small
\resizebox{\columnwidth}{!}{%
\begin{tabular}{lccc}
\toprule
\textbf{Score} & \textbf{Err.\ enrich.$\times$} & \textbf{AUROC} & \textbf{AURC} \\
\midrule
Neg. candidate entropy & 3.50$\pm$1.07 & 0.836$\pm$0.143 & 0.0189$\pm$0.0179 \\
Crop agreement & 3.83$\pm$0.99 & 0.772$\pm$0.095 & 0.0222$\pm$0.0215 \\
Max probability & 2.77$\pm$1.72 & 0.782$\pm$0.172 & 0.0224$\pm$0.0201 \\
Full composite & 2.60$\pm$1.92 & 0.728$\pm$0.117 & 0.0249$\pm$0.0209 \\
Energy margin & 2.37$\pm$2.05 & 0.692$\pm$0.134 & 0.0270$\pm$0.0221 \\
Context confidence & 0.67$\pm$1.15 & 0.542$\pm$0.105 & 0.0439$\pm$0.0317 \\
\bottomrule
\end{tabular}%
}
\end{table}

\section{Supplementary: Per-Code Accuracy}

Table~\ref{tab:percode} shows AURA-C exact match broken down by active code (3-bit antibiotic combination). Codes with fewer images show higher variance; AURA maintains high accuracy across all well-represented codes.

\begin{table}[h]
\centering
\caption{AURA-C per-code accuracy on pooled ResNet-18 main-split test images. Code format is cipro--cef--genta, where 1 indicates active response. The rare 110 code remains the most difficult case.}
\label{tab:percode}
\setlength{\tabcolsep}{4pt}
\small
\begin{tabular}{llcc}
\toprule
\textbf{Code} & \textbf{Interpretation} & \textbf{$n$} & \textbf{Acc.\ (\%)} \\
\midrule
000 & None active & 169 & 98.22 \\
001 & Genta only & 4 & 100.00 \\
010 & Cef only & 64 & 100.00 \\
011 & Cef+Genta & 15 & 86.67 \\
100 & Cipro only & 10 & 100.00 \\
101 & Cipro+Genta & 18 & 100.00 \\
110 & Cipro+Cef & 23 & 52.17 \\
111 & All active & 20 & 95.00 \\
\bottomrule
\end{tabular}
\end{table}

\section{Supplementary: DINOv2 and ViT Detailed Results}

Table~\ref{tab:dino_detail} shows detailed DINOv2 results for AURA-C and inverse baselines.

\begin{table}[h]
\centering
\caption{DINOv2-ViT-S/14 robustness results on the three main splits. AURA-C remains ahead of the strongest non-AURA context-rule baseline.}
\label{tab:dino_detail}
\setlength{\tabcolsep}{3pt}
\small
\resizebox{\columnwidth}{!}{%
\begin{tabular}{lccc}
\toprule
\textbf{Method} & \textbf{Exact (\%)} & \textbf{Macro-F1} & \textbf{Hamming} \\
\midrule
AURA-C & 95.67$\pm$2.09 & 94.40$\pm$4.29 & 97.67$\pm$0.96 \\
AURA w/o context & 96.11$\pm$3.53 & 94.87$\pm$4.56 & 97.97$\pm$1.80 \\
AURA oracle$^\dagger$ & 99.01$\pm$0.86 & 98.42$\pm$1.59 & 99.34$\pm$0.58 \\
Context LR(img+applied)$\to$rule & 87.10$\pm$9.55 & 86.11$\pm$12.72 & 94.36$\pm$4.34 \\
Ind. LR + vote & 83.04$\pm$5.22 & 81.66$\pm$2.07 & 91.93$\pm$1.81 \\
Latent/sparse/dict. inv. & 72.84$\pm$15.04 & 68.55$\pm$14.76 & 87.40$\pm$6.96 \\
Nearest prototype & 76.74$\pm$13.49 & 78.27$\pm$8.25 & 90.02$\pm$5.31 \\
scGen VAE inverse & 75.14$\pm$13.92 & 77.65$\pm$6.43 & 89.35$\pm$5.44 \\
\bottomrule
\end{tabular}%
}
\end{table}

\begin{table}[h]
\centering
\caption{ViT-B/16 robustness results on the three main splits. AURA-C remains above the best non-AURA context-rule baseline, although the margin is smaller than with ResNet-18 and DINOv2.}
\label{tab:vit_detail}
\setlength{\tabcolsep}{3pt}
\small
\resizebox{\columnwidth}{!}{%
\begin{tabular}{lccc}
\toprule
\textbf{Method} & \textbf{Exact (\%)} & \textbf{Macro-F1} & \textbf{Hamming} \\
\midrule
AURA-C & 92.20$\pm$3.26 & 91.91$\pm$3.99 & 96.50$\pm$0.32 \\
AURA w/o context & 92.20$\pm$3.26 & 91.72$\pm$3.83 & 96.42$\pm$0.23 \\
AURA oracle$^\dagger$ & 94.64$\pm$5.42 & 95.49$\pm$2.56 & 97.97$\pm$1.59 \\
Context LR(img+applied)$\to$rule & 90.47$\pm$3.28 & 87.88$\pm$8.64 & 94.82$\pm$1.72 \\
Ind. LR + vote & 82.26$\pm$8.21 & 81.57$\pm$5.77 & 91.82$\pm$2.91 \\
Latent/sparse/dict. inv. & 73.71$\pm$14.08 & 67.96$\pm$11.77 & 87.80$\pm$6.14 \\
Nearest prototype & 72.92$\pm$16.70 & 72.22$\pm$8.36 & 88.19$\pm$5.96 \\
scGen VAE inverse & 75.49$\pm$15.85 & 78.25$\pm$5.36 & 90.03$\pm$4.99 \\
\bottomrule
\end{tabular}%
}
\end{table}

\section{Crop Robustness test}
Because saliency maps on the original segmented crops were not sufficiently informative, we performed an object-focused crop robustness control. Each segmented crop was recentered on the non-background bacterial object with a reduced 160$\times$160 canvas. AURA-C remained stable under this perturbation, changing from 95.47\% to 95.01\% exact match and from 93.39\% to 93.80\% macro-F1. This suggests that the model is not primarily exploiting black-canvas or crop-padding artifacts, but retains the active-response signal from the bacterial object itself.
\begin{table}[t]
\centering
\caption{Object-focused crop robustness. Original uses the standard 256$\times$256 segmented crop canvas. Object-focused crops recenter the non-background bacterial object on a smaller 160$\times$160 canvas. AURA-C remains stable, indicating that the result is not driven by crop canvas or padding artifacts.}
\label{tab:object_crop_robustness}
\small
\resizebox{\columnwidth}{!}{%
\begin{tabular}{lcccccc}
\toprule
Method & Orig. Exact & Obj. Exact & $\Delta$ Exact & Orig. F1 & Obj. F1 & $\Delta$ F1 \\
\midrule
AURA-C & 95.47 & 95.01 & -0.46 & 93.39 & 93.80 & +0.40 \\
AURA w/o context & 95.21 & 94.75 & -0.46 & 93.54 & 92.81 & -0.73 \\
Nearest prototype & 74.23 & 74.42 & +0.20 & 71.28 & 72.51 & +1.23 \\
Latent arithmetic inverse & 74.47 & 73.49 & -0.98 & 73.20 & 70.52 & -2.68 \\
\bottomrule
\end{tabular}
}
\end{table}

Under the object-focused crop setting, AURA-C still exceeds the strongest context-rule baseline by 3.38 percentage points, preserving the main ranking observed with the original crop canvas.

\section{Supplementary: CPA/scGen/IMPA Detailed Results}

\begin{table}[h]
\centering
\caption{Paper-manual CPA, scGen, and IMPA inverse results on the ResNet-18 main splits. ``Known-context'' rows use context information inside the corresponding forward-model inverse wrapper. They are diagnostic compatibility checks rather than fair AURA competitors, because AURA's main test-time setting does not provide the true strain/context label.}
\label{tab:cpa_scgen}
\setlength{\tabcolsep}{3pt}
\small
\resizebox{\columnwidth}{!}{%
\begin{tabular}{lcccc}
\toprule
\textbf{Method} & \textbf{Exact} & \textbf{Macro-F1} & \textbf{Macro Bal.\ Acc.} & \textbf{Hamming} \\
\midrule
\multicolumn{5}{l}{\textit{CPA}} \\
CPA covariate-enumerated inverse & 30.82$\pm$20.86 & 34.75$\pm$13.31 & 59.23$\pm$4.01 & 71.00$\pm$8.58 \\
CPA known-context inverse & 31.02$\pm$19.94 & 33.46$\pm$16.21 & 58.85$\pm$5.01 & 70.97$\pm$7.87 \\
\midrule
\multicolumn{5}{l}{\textit{scGen}} \\
scGen VAE latent-delta inverse & 67.41$\pm$19.26 & 68.51$\pm$5.10 & 77.40$\pm$3.63 & 84.33$\pm$7.97 \\
scGen linear feature inverse & 73.06$\pm$12.00 & 68.94$\pm$5.45 & 77.44$\pm$3.29 & 86.87$\pm$5.11 \\
scGen PCA vector inverse & 74.44$\pm$12.88 & 68.06$\pm$2.93 & 76.85$\pm$0.32 & 87.31$\pm$4.83 \\
scGen known-context inverse & 71.90$\pm$16.97 & 74.39$\pm$7.14 & 81.19$\pm$3.89 & 87.00$\pm$7.60 \\
\midrule
\multicolumn{5}{l}{\textit{IMPA}} \\
IMPA generation-feature inverse & 64.25$\pm$16.79 & 61.94$\pm$10.38 & 75.08$\pm$5.85 & 84.43$\pm$6.61 \\
IMPA multitask-discriminator inverse & 41.61$\pm$17.91 & 53.72$\pm$1.90 & 70.26$\pm$6.55 & 72.07$\pm$8.44 \\
\bottomrule
\end{tabular}%
}
\end{table}

\section{Supplementary: Diagnostic Analysis of the 110 Code}

Table~\ref{tab:110diag} summarizes the diagnostic analysis for the weakest active-response code, 110 (ciprofloxacin+ceftriaxone). The single-source settings contain more crop-level samples than field-level samples, but crops from the same microscopy field are correlated and therefore do not represent independent biological fields.

\begin{table}[h]
\centering
\caption{Diagnostic support and performance for active-response code 110. The key limitation is low independent field-level support and no-response-like morphology for a subset of 110 fields, rather than arbitrary label confusion.}
\label{tab:110diag}
\setlength{\tabcolsep}{4pt}
\small
\resizebox{\columnwidth}{!}{%
\begin{tabular}{lccccc}
\toprule
\textbf{Split} & \textbf{Train fields} & \textbf{Train crops} & \textbf{Test fields} & \textbf{Test crops} & \textbf{110 acc.} \\
\midrule
D1$\to$D2 & 7 & 21 & 9 & 51 & 22.22 \\
D2$\to$D1 & 9 & 51 & 7 & 21 & 57.14 \\
D1+D2$\to$D3 & 16 & 72 & 7 & 84 & 85.71 \\
\bottomrule
\end{tabular}
}
\end{table}

The pooled confusion pattern for true 110 is structured: 12 fields are correctly predicted as 110, 8 are predicted as 000, and 3 are predicted as 010. Descriptor-centroid analysis further shows that 60.87\% of 110 fields are nearest to the 110 centroid and 39.13\% are nearest to the 000 centroid, while 0.00\% are nearest to the 010 centroid. Descriptor-only classification also separates 110 from 010 strongly (AUC = 1.000, balanced accuracy = 0.992). These diagnostics indicate that the main difficulty is a subset of 110 fields appearing no-response-like, not broad morphological overlap with cef-only response.

\section{Supplementary: Coefficient-Derived Sparse Inverse Baselines}

The main table collapses ElasticNet, NNLS, latent arithmetic, and ridge-dictionary inverse baselines because their reconstruction-ranked candidate decisions are identical on the ResNet-18 main splits. To verify that this collapse does not hide a stronger sparse solver, we additionally evaluated coefficient-derived NNLS and ElasticNet variants. In these variants, active bits are obtained directly by thresholding fitted antibiotic-response coefficients, rather than by selecting the candidate subset with the lowest reconstruction energy.

\begin{table}[h]
\centering
\caption{Coefficient-derived sparse inverse baselines on the ResNet-18 main splits. Source-selected thresholds are chosen using source/train images only, not target labels. These variants improve over fixed-threshold sparse coding but remain below AURA-C and the context-rule baseline.}
\label{tab:coeff_sparse}
\setlength{\tabcolsep}{4pt}
\small
\resizebox{\columnwidth}{!}{%
\begin{tabular}{lcc}
\toprule
\textbf{Method} & \textbf{Exact} & \textbf{Macro-F1} \\
\midrule
ElasticNet coeff. threshold, source-selected & 79.12$\pm$6.97 & 77.75$\pm$4.48 \\
NNLS coeff. threshold, source-selected & 78.85$\pm$6.70 & 77.47$\pm$4.95 \\
ElasticNet coeff. threshold, fixed 0.10 & 69.23$\pm$1.67 & 76.24$\pm$7.93 \\
NNLS coeff. threshold, fixed 0.10 & 68.77$\pm$2.17 & 75.84$\pm$8.17 \\
\bottomrule
\end{tabular}
}
\end{table}

The best coefficient-derived sparse variant reaches 79.12\% exact match, which improves over the reconstruction-ranked collapsed inverse row but remains substantially below AURA-C (95.47\%) and the strongest context-rule baseline (92.39\%). Thus, the collapse of the reconstruction-ranked inverse baselines does not affect the main conclusion.

\begin{rxblock}
\section{Supplementary: Frozen RxRx3 Pseudo-Cocktail Results}

\subsection{Protocol and Main Summary}
The frozen RxRx3 protocol uses fixed well-level morphology embeddings as inputs. Source-split embeddings are used to estimate context baselines and perturbation response atoms; target examples are then formed as a real held-out target embedding for $p_1$ plus a source-derived atom for $p_2$. Candidate sets contain the active pair and hard morphology distractors. Table~\ref{tab:rxrx3_splitwise} provides split-wise exact-pair accuracy, and Table~\ref{tab:rxrx3_splitk} shows how AURA-C and the strongest non-AURA baseline vary with candidate-set size.

\begin{table*}[t]
\centering
\color{\rxcolor}
\caption{Split-wise RxRx3 exact-pair accuracy (\%) for the frozen target-plus-atom protocol. Values are averaged over $K=8,16,32$ and seeds 44--46.}
\label{tab:rxrx3_splitwise}
\setlength{\tabcolsep}{2pt}
\small
\resizebox{\textwidth}{!}{%
\begin{tabular}{lccc}
\toprule
\textbf{Method} & \textbf{D1$\to$D2} & \textbf{D2$\to$D1} & \textbf{D1+D2$\to$D3} \\
\midrule
\textbf{AURA-C trained ($\lambda=0.25$)} & \textbf{65.78} & \textbf{62.10} & \textbf{54.91} \\
NNLS coeff. top-2 & 24.58 & 25.01 & 23.16 \\
Ridge dictionary pair energy & 14.77 & 13.29 & 16.13 \\
ElasticNet positive top-2 & 14.73 & 14.69 & 14.64 \\
AURA empirical energy & 11.59 & 11.59 & 12.76 \\
scGen latent-shift inverse & 9.14 & 9.73 & 10.00 \\
Candidate pair prototype & 5.21 & 5.19 & 5.52 \\
CPA additive inverse & 4.70 & 4.38 & 4.62 \\
IMPA nonlinear forward inverse & 4.22 & 4.48 & 4.71 \\
Candidate LR top-2 scorer & 4.63 & 4.18 & 4.54 \\
Candidate nearest top-2 prototype & 3.44 & 3.64 & 4.14 \\
Random top-2 & 1.79 & 1.54 & 1.58 \\
\bottomrule
\end{tabular}%
}
\end{table*}

\begin{table*}[t]
\centering
\color{\rxcolor}
\caption{RxRx3 split-by-candidate-size exact-pair accuracy (\%) for AURA-C and the strongest non-AURA baseline. Values are averaged over three held-out seeds.}
\label{tab:rxrx3_splitk}
\setlength{\tabcolsep}{2.5pt}
\scriptsize
\resizebox{\textwidth}{!}{%
\begin{tabular}{lccccccccc}
\toprule
\textbf{Method} & \textbf{D1$\to$D2 K8} & \textbf{D1$\to$D2 K16} & \textbf{D1$\to$D2 K32} & \textbf{D2$\to$D1 K8} & \textbf{D2$\to$D1 K16} & \textbf{D2$\to$D1 K32} & \textbf{D1+D2$\to$D3 K8} & \textbf{D1+D2$\to$D3 K16} & \textbf{D1+D2$\to$D3 K32} \\
\midrule
\textbf{AURA-C trained ($\lambda=0.25$)} & \textbf{73.03} & \textbf{67.43} & \textbf{56.87} & \textbf{74.30} & \textbf{64.67} & \textbf{47.33} & \textbf{64.90} & \textbf{58.80} & \textbf{41.03} \\
NNLS coeff. top-2 & 31.83 & 27.03 & 14.87 & 34.47 & 28.00 & 12.57 & 31.37 & 25.30 & 12.80 \\
\bottomrule
\end{tabular}%
}
\end{table*}

\subsection{Paired Statistical Tests}
Table~\ref{tab:rxrx3_paired} reports prediction-level paired tests comparing AURA-C with every non-AURA baseline over the same 27,000 cases. All comparisons remain significant after Holm correction.

\begin{table*}[t]
\centering
\color{\rxcolor}
\caption{Prediction-level paired tests on the frozen RxRx3 target-plus-atom protocol. Delta is AURA-C exact-pair accuracy minus the baseline exact-pair accuracy in percentage points.}
\label{tab:rxrx3_paired}
\setlength{\tabcolsep}{4pt}
\small
\resizebox{\textwidth}{!}{%
\begin{tabular}{lcccc}
\toprule
\textbf{Baseline} & \textbf{Baseline Exact (\%)} & \textbf{$\Delta$ Exact (pp)} & \textbf{95\% CI (pp)} & \textbf{Holm $p$} \\
\midrule
NNLS coeff. top-2 & 24.25 & 36.68 & [36.02, 37.34] & 0.0011 \\
Ridge dictionary pair energy & 14.73 & 46.20 & [45.54, 46.89] & 0.0011 \\
ElasticNet positive top-2 & 14.69 & 46.24 & [45.57, 46.93] & 0.0011 \\
AURA empirical energy & 11.98 & 48.95 & [48.30, 49.63] & 0.0011 \\
scGen latent-shift inverse & 9.63 & 51.30 & [50.65, 51.98] & 0.0011 \\
Candidate pair prototype & 5.31 & 55.62 & [55.00, 56.24] & 0.0011 \\
CPA additive inverse & 4.57 & 56.36 & [55.76, 56.97] & 0.0011 \\
IMPA nonlinear forward inverse & 4.47 & 56.46 & [55.85, 57.08] & 0.0011 \\
Candidate LR top-2 scorer & 4.45 & 56.48 & [55.87, 57.09] & 0.0011 \\
Candidate nearest top-2 prototype & 3.74 & 57.19 & [56.58, 57.80] & 0.0011 \\
Random top-2 & 1.64 & 59.29 & [58.71, 59.89] & 0.0011 \\
\bottomrule
\end{tabular}%
}
\end{table*}

\subsection{Ablations and AURA-E}
The RxRx3 ablation and AURA-E results are integrated into the main combined ablation and selective-prediction tables (Tables~\ref{tab:ablation} and~\ref{tab:aurae}). Paired RxRx3 ablation tests confirm the same pattern: the reference AURA-C model improves over the no-candidate-set variant by 57.17 percentage points (95\% CI [56.58, 57.76], Holm $p=0.0004$), while context-prior, class-balance, and margin-loss ablations differ by less than 0.6 percentage points.

\subsection{Component Audit: Base $p_1$ vs Added $p_2$}
Because the pseudo-cocktail contains a real held-out sample for $p_1$, Table~\ref{tab:rxrx3_component} separates exact-pair recovery from base-$p_1$ and added-$p_2$ hit rates. AURA-C outperforms NNLS on exact-pair recovery and also improves added-$p_2$ recovery, supporting that the gain is not solely due to recognizing the visible base component.

\begin{table*}[t]
\centering
\color{\rxcolor}
\caption{RxRx3 component audit. Base-$p_1$ hit indicates whether the method selected the real target-sample perturbation; added-$p_2$ hit indicates whether it selected the injected response component.}
\label{tab:rxrx3_component}
\setlength{\tabcolsep}{4pt}
\small
\resizebox{\textwidth}{!}{%
\begin{tabular}{lcccc}
\toprule
\textbf{Method} & \textbf{$n$} & \textbf{Exact Pair (\%)} & \textbf{Base $p_1$ Hit (\%)} & \textbf{Added $p_2$ Hit (\%)} \\
\midrule
\textbf{AURA-C trained ($\lambda=0.25$)} & 27000 & \textbf{60.93} & \textbf{68.40} & \textbf{71.86} \\
NNLS coeff. top-2 & 27000 & 24.25 & 35.88 & 62.17 \\
Ridge dictionary pair energy & 27000 & 14.73 & 19.49 & 59.33 \\
ElasticNet positive top-2 & 27000 & 14.69 & 27.60 & 64.40 \\
AURA empirical energy & 27000 & 11.98 & 17.37 & 62.76 \\
scGen latent-shift inverse & 27000 & 9.63 & 15.05 & 67.95 \\
Candidate pair prototype & 27000 & 5.31 & 17.25 & 52.10 \\
CPA additive inverse & 27000 & 4.57 & 16.07 & 54.23 \\
IMPA nonlinear forward inverse & 27000 & 4.47 & 16.77 & 42.46 \\
Candidate LR top-2 scorer & 27000 & 4.45 & 15.10 & 58.51 \\
Candidate nearest top-2 prototype & 27000 & 3.74 & 15.29 & 46.67 \\
Random top-2 & 27000 & 1.64 & 14.73 & 14.89 \\
\bottomrule
\end{tabular}%
}
\end{table*}

A separate context-shifted pseudo-cocktail branch is ongoing and currently shows a comparable AURA-vs-best-non-AURA margin, but we do not use it as the primary public stress-test claim because the frozen target-plus-atom protocol is complete, audited, and simpler to interpret.
\end{rxblock}

\end{document}